\documentclass[a4paper]{article}
\usepackage{algorithmic}
\usepackage{algorithm}
\usepackage{amsmath}
\usepackage{graphicx}
\usepackage{wrapfig}
\usepackage{amsfonts}

\usepackage{times}  
\usepackage{helvet}  
\usepackage{courier}  
\usepackage[hyphens]{url}  
\usepackage{graphicx} 
\usepackage{natbib}  
\usepackage{caption} 
\title{On the complexity of PAC learning in Hilbert spaces\footnote{To appear in the proceedings of the AAAI-2023 conference.}}

\author{Sergei Chubanov\footnote{{sergei.chubanov@de.bosch.com} Bosch Center for Artificial Intelligence, Renningen, Germany.}}

\newtheorem{definition}{Definition}

\newtheorem{lemma}{Lemma}
\newtheorem{theorem}{Theorem}

\newtheorem{corollary}{Corollary}
\newtheorem{condition}{Condition}

\def\BlackBox{\hspace*{\fill}\vrule width 4pt height 4pt depth 0pt}

\newcommand{\inprod}[2]{\langle #1,#2 \rangle}

\newcommand{\samplecomplexity}[0]{O(\varepsilon^{-1}D + \varepsilon^{-1}\log(\delta^{-1}))}
\newcommand{\vcdimension}[0]{O((\gamma^{-2} + \log t)(t\gamma^{-2})^{4\gamma^{-2}})}

\newcommand{\map}{\longrightarrow}

\begin{document}

\maketitle

\begin{abstract}
We study the problem of binary classification from the point of view of learning convex polyhedra in Hilbert spaces, to which one can reduce any binary classification problem. The problem of learning convex polyhedra in finite-dimensional spaces is sufficiently well studied in the literature. We generalize this problem to that in a Hilbert space and propose an algorithm for learning a polyhedron which correctly classifies at least $1- \varepsilon$ of the distribution, with a probability of at least $1 - \delta,$ where $\varepsilon$ and $\delta$ are given parameters. Also, as a corollary, we improve some previous bounds for polyhedral classification in finite-dimensional spaces. 

\end{abstract}

\section{Introduction}\label{sec:rel_work}


In general, the classification problem we are dealing with  can be formulated as follows: Find a binary classifier which
consistently classifies the training data such that the number of elementary operations involved in the description of the classifier is as small as possible. The intuition behind restricting the complexity of the classifier is based on the well-known Occam's razor principle; it has been proved (see e.g. \cite{DBLP:journals/ipl/BlumerEHW87, DBLP:journals/jacm/BlumerEHW89}) that there is a relationship between the complexity of a classifier and its prediction accuracy, in the sense of the probably approximately correct (PAC) classification. 


We study this problem from the point of view of polyhedral classification, where the concept class to be learned consists of polyhedra in a given inner-product space. Polyhedral separability is always realizable by choosing a suitable kernel; i.e., our results are universally applicable to the general case of binary classification. 

One should note that, unless $P = NP,$ we cannot hope for a polynomial-time method for finding a polyhedral classifier defined by the minimum possible number of halfspaces because even the problem of polyhedral separation in finite-dimensional spaces by means of two halfspaces is NP-hard \cite{Megiddo1996}.

{\bf Our results and related work.} We propose an algorithm for both {\em proper} and {\em improper} learning of polyhedral concepts in inner-product spaces that can be finite-dimensional Euclidean spaces or infinite-dimensional Hilbert spaces. In the context of our paper, proper learning means learning a polyhedron defined by $t$ halfspaces ($t$-polyhedron), under the assumption that there exists a $t$-polyhedron consistently classifying the entire instance space. Similarly to the previous publications with which we compare our results, we assume that the halfspaces defining a hypothetical ground-truth polyhedron do not contain any instance of the instance space in their $\gamma$-margins, for some $\gamma > 0.$  

Consistent polyhedral separation is a well-known problem which has been proved to be intractable even in the case $t = 2,$ as we have already mentioned. Despite to the NP-hardness of exact separation with a given number of inequalities in finite-dimensional spaces, there are different non-trivial results concerning exponential bounds on the running time of learning algorithms; see \cite{DBLP:conf/nips/GottliebKKN18}. Our contribution to this line of research is the following: First, we propose a new algorithm allowing to improve existing bounds on the running time of PAC learning algorithms for the finite-dimensional case. Second, our algorithm is applicable to infinite-dimensional Hilbert spaces, in which case it guarantees similar complexity bounds in terms of a model of computation including inner products as elementary operations. 



To the best of our knowledge, the current tightest bounds for both proper and improper learning have been obtained in \cite{DBLP:conf/nips/GottliebKKN18}. There, the authors proposed algorithms for both proper and improper learning of polyhedral concepts in finite-dimensional spaces. 
A follow-up work related to \cite{DBLP:conf/nips/GottliebKKN18} appeared recently in \cite{Gottlieb2022LearningCP} with some corrections of the previous work. For a more exact comparison, we give a brief summary of their performance bounds compared to ours:
\begin{itemize}
\item \cite{Gottlieb2022LearningCP}: 
\begin{itemize} 
\item Running time for proper classification: \[m^{O(t\gamma^{-2}\log(\gamma^{-1}))},\] where $m$ is the sample complexity.
\item Running time for improper classification: \[m^{O(\gamma^{-2}\log(\gamma^{-1}))},\]
\item Sample complexity: \[m = O(D{\varepsilon^{-1}}\log^2(\varepsilon^{-1}) + \log(\delta^{-1})),\] where $D = (\gamma^{-2}t\log t).$\\ 
\item Applicability to Hilbert spaces: In its present state, that approach is not applicable to infinite-dimensional spaces because some of its key components like the so-called fat-shuttering dimension analyzed in Lemma 2.2 of \cite{Gottlieb2022LearningCP} depend on the dimension of the vector space in question.
\end{itemize}
\item Ours (with respect to a finite-dimensional space): 
\begin{itemize} 
\item Running time for proper classification: $m^{O(t\gamma^{-2}).}$
\item Running time for improper classification: \[O\left(d\cdot m\cdot(t\gamma^{-2})^{4\gamma^{-2}}\right).\]
\item Sample complexity $m$: 
\begin{itemize}
\item Proper learning: $O(dt\log t),$ where $d$ is the dimension of the space.
\item Improper learning: $\samplecomplexity$ where $D = \vcdimension.$
\end{itemize}
\end{itemize}
\end{itemize}
For proper classification, we save $\log(\gamma^{-1})$ in the power of the exponent, compared to \cite{Gottlieb2022LearningCP}. For improper classification, the running time of our algorithm is substantially better, but the sample complexity is exponential in $\gamma^{-2}.$

An improper learning algorithm for finite-dimensional spaces with the running time depending exponentially on the dimension of the space and without a margin assumption, but with additional assumptions about the probability distribution, was presented in \cite{vempala2010} and \cite{klivans-donnell-servedio-rocco}.

Support vector machines~\cite{conf/icml/BennettB00,DBLP:journals/tnn/MavroforakisT06} as large margin classifiers are certainly also related to our work. In particular, in~\cite{conf/icml/BennettB00} a relevant geometric interpretation of SVM has been proposed. Beyond the kernel methods like SVM, research on large-margin classification directly in Hilbert~\cite{DBLP:journals/ijon/RossiV06} and Banach~\cite{DBLP:journals/jmlr/DerL07} spaces was initiated within the area of functional data analysis. 

The problem of polyhedral classification in Euclidean spaces is known under various names, e.g., learning convex polytopes~\cite{DBLP:conf/nips/GottliebKKN18}, linear decision lists~\cite{DBLP:journals/dm/Jeroslow75,DBLP:journals/jmlr/Anthony04,DBLP:journals/tit/Mangasarian68}, neural lists~\cite{DBLP:conf/nips/LongS06}, or threshold functions~\cite{DBLP:journals/dam/AnthonyR12}. Many of the methods for learning such structures (see \cite{DBLP:journals/jmlr/Anthony04} for overview) can be viewed as cutting-plane methods where inconsistencies are resolved by sequentially adding and reorganizing new halfspaces or hyperplanes at each iteration. These methods have several variations including a multisurface method proposed in~\cite{DBLP:journals/tit/Mangasarian68,Mangasarian90patternrecognition}, where at each stage two parallel hyperplanes located close to each other are identified, such that the points not enclosed between them have the same classification.

In ~\cite{DBLP:journals/jmlr/ManwaniS10}, the authors propose an approach for learning polyhedral classifiers based on logistic function. 
Paper ~\cite{NIPS2014_5511} proposes an algorithm for learning polyhedral concepts using stochastic gradient descent.
The work~\cite{DBLP:journals/jmlr/Anthony04} analyses generalization properties of techniques based on the use of linear decision lists and presents a bound on the generalization error in a standard probabilistic learning framework. The authors of~\cite{DBLP:journals/corr/abs-1901-05911} have recently demonstrated a lower-bound technique for linear decision lists, which has been used to prove an explicit lower bound. 


\section{Preliminaries}

In a Hilbert space $\mathcal{H,}$ a halfspace can be represented as $\{z\in\mathcal{H}:\inprod{h}{z} + d \ge 0\}$ for some $h\in\mathcal{H}$ and $d\in\mathbb{R}.$ Here, $\inprod{\cdot}{\cdot}$ is the inner product. Since it will always be known from the context what spaces the respective elements belong to, we will not use an additional notation to specify the space with which the respective inner product is associated. For instance, we will write $\inprod{(h_1, d_1)}{(h_2, d_2)}$ for the inner product of elements $(h_i, d_i),$ $i = 1,2,$ of the Cartesian product $\mathcal{H}\times\mathbb{R},$ where $h_i\in\mathcal{H}$ and $d_i\in\mathbb{R},$ meaning that $\inprod{(h_1, d_1)}{(h_2, d_2)} = \inprod{h_1}{h_2} + d_1d_2.$

{\bf Large margins and polyhedra.} Regarding large-margin polyhedral classification, we will partially rely on the terminology introduced in some previous works on polyhedral classification in Euclidean spaces. The outer $\gamma$-margin of a halfspace $H$ is the set of points not in $H$ whose distance to its boundary $\partial H$ is not greater than $\gamma.$ The inner $\gamma$-margin of $H$ is the set of points in $H$ whose distance to $\partial H$ is not greater than $\gamma.$ The $\gamma$-margin of $H$ is the union of its inner and outer $\gamma$-margin. We denote by $H^\gamma$ the union of $H$ and its $\gamma$-margin.

We say that two sets $X^-$ and $X^+$ are linearly $\gamma$-separable if there is a halfspace $H$ such that $H\cap X^- = \emptyset$ and $X^+\subset H$ and the $\gamma$-margin of $H$ does not contain any point of $X^-\cup X^+.$ For short, we will equivalently say that $X$ is linearly $\gamma$-separable.

A polyhedron $F$ is the intersection $\cap_\alpha H_\alpha$ of a finite collection of some halfspaces $H_\alpha.$ Let $F^\gamma$ be the polyhedron $\cap_\alpha H^\gamma_\alpha.$
The $\gamma$-margin of $F$ is $F^\gamma\setminus F.$ If for some set $X$ polyhedron $F$ contains $X^+$ and does not intersect with $X^-,$ and at the same time the $\gamma$-margin of $F$ does not intersect with the whole set $X,$ we say that $F$ $\gamma$-separates $X^-$ from $X^+.$ A $t$-polyhedron is a polyhedron being an intersection of $t$ halfspaces.

{\bf Instance space.} The instance space $\mathcal{X}\subset\mathcal{H}$ is assumed to be contained in a unit ball centered at the origin ${\bf 0}$ of $\mathcal{H}.$ That is, $\|x\| \le 1$ for all $x\in\mathcal{X}.$
We assume that there are two class labels $-1$ and $1$ and that the instance space $\mathcal{X}$ is equipped with a function $y:\mathcal{X}\longrightarrow \{-1, 1\}$ with the property that there exists $\rho > 0$ such that 
\begin{equation}\label{rho-inequality}
 \forall x, x^\prime \in \mathcal{X}: y(x) \ne y(x^\prime) \Longrightarrow \|x - x^\prime\| \ge \rho.
\end{equation}
This condition tells that two instances of different classes cannot be arbitrarily close to each other. That is, the distance between $\mathcal{X}^-$ and $\mathcal{X}^+$ is nonzero, where $\mathcal{X}^-$ and $\mathcal{X}^+$ are the respective classes.

Further, when considering a subset $X$ of $\mathcal{X},$ we will denote by $X^-$ the set of all negatives, i.e., those $x\in X$ with $y(x) = -1$ and by $X^+$ the set of all positives in $X,$ i.e., those with $y(x) = 1.$

{\bf Machine-learning framework.} When talking about probability distributions, we assume that our space is a certain reproducing kernel Hilbert space of a sufficiently well-behaved kernel over an underlying Euclidean space such that our instance space in the Hilbert space is an image of a compact subset $\Omega$ of that Euclidean space. Drawing a sample (a finite subset of instances in $\mathcal{X}$) means drawing a sample from a fixed probability distribution defined on $\Omega$ and then considering its image in the Hilbert space. 

We say that a subset $H$ of $\mathcal{H}$ correctly classifies $x\in\mathcal{X}$ if $y(x) = 1$ implies that $x\in H$ and $y(x) = -1$ implies that $x\not\in H.$ In this role, we often refer to $H$ as a classifier.

To estimate the prediction accuracy of polyhedral classifiers constructed by our algorithms, we use the classical framework of probably approximately correct (PAC) learning based on VC dimension; $1- \varepsilon$ and $1 - \delta$ are the prediction accuracy and the confidence estimate, respectively, where $\varepsilon$ and $\delta$ are given values in $(0,1/2).$ For the finite-dimensional case, the VC dimension of linear concepts is bounded by the dimension $d$ of the space. The VC dimension of the family of polyhedra defined by $t$ halfspaces is bounded by $d\cdot t\log t,$ which is a well-known fact. The infinite-dimensional case is more difficult, although we can also provide a similar estimate where in place of $d$ we use $O(\gamma^{-2})$ under the assumption that there is a $t$-polyhedron that $\gamma$-separates the instance space. However, our algorithmic approach, though working for the finite-dimensional case, does not guarantee that the consistent polyhedron learned by the algorithm belongs to this concept class. To overcome this difficulty, we use discretization and an additional assumption that the Hilbert space in question is a reproducing kernel Hilbert space (RKHS) with a sufficiently well-behaved kernel. The discretization allows us to learn from a finite concept class. 

{\bf Realizability.} Polyhedral separability of the image of any finite-dimensional data with binary labels is always realizable in a suitable RKHS. E.g., we can choose an RBF kernel $K$ or a suitable function $K$ of a given kernel $K^\prime$ to guarantee that $K(x, x) = 1, K(x, x^\prime) < 1,$ for all $x, x^\prime\in\mathcal{X},$ $x\ne x^\prime.$ Then we can prove that the image of our original $\mathcal{X}$ in the RKHS is $\gamma$-separable by a $t$-polyhedron, for some $t$ and $\gamma,$ because of (\ref{rho-inequality}); see the appendix for details.

{\bf Model of computation.} The elementary operations are computations of inner products and norms in the vector spaces involved, and arithmetic operations. 


%

\section{Linear separation in Hilbert spaces}\label{sec:intro}

In this section, we consider the following system $P(X)$ of strict linear inequalities associated with a subset $X$ of $\mathcal{X}:$
\[
P(X): y(x)(\inprod{x}{h} + d) > 0, \forall x\in X.
\]
where we are looking for $(h, d)\in\mathcal{H}\times\mathbb{R}.$ 
A pair $(h, d)$ defines a halfspace \[H = \{z\in\mathcal{H}: \inprod{z}{h} + d \ge 0\},\] which can serve as a linear classifier; If $x\in H,$ it assigns label $1$ to $x,$ otherwise, label $-1$ is assigned. If $(h, d)$ is feasible for $P(X),$ then $H$ assigns correct labels for all $x\in X.$


\begin{algorithm}[tb]
\caption{\bf Linear-programming (LP) algorithm}
\label{LP-algorithm}
\begin{algorithmic}
\STATE{{\bf Input:} System $P(X),$ state $S$ compatible with $X,$ and $\gamma.$}
\STATE{{\bf Output:} State $S^\prime.$}
\STATE{$S^\prime := S.$}
\WHILE{$S^\prime = \emptyset$ or $(h_{S^\prime}, d_{S^\prime})$ is not feasible for $P(X)$ and $\|(h_{S^\prime}, d_{S^\prime})\| \ge \gamma/2$}
\IF{$S^\prime\ne\emptyset$}
\STATE{(Progress-contributing step)}
\STATE{$(h, d) := (h_{S^\prime}, d_{S^\prime})$}
\STATE{Find $x\in X$ with $y(x)(\inprod{x}{h_{S^\prime}} + d_{S^\prime}) \le 0.$} 
\STATE{Let $(h^\prime, d^\prime)$ be the orthogonal projection of $({\bf 0}, 0)$ onto $[(h,d), y(x)(x, 1)]$ and $S^\prime$ be defined by $(h^\prime, d^\prime).$}
\ELSE
\STATE{Pick an arbitrary $x\in X.$}
\STATE{$(h^\prime, d^\prime) := y(x)(x, 1)$}
\STATE{Let $S^\prime$ be defined by $(h^\prime, d^\prime).$}
\ENDIF
\IF{ $\|(h_{S^\prime}, d_{S^\prime})\| < \gamma/2$}
\STATE{Report that $X$ is not linearly $\gamma$-separable.}
\ENDIF
\ENDWHILE
\STATE{Return $S^\prime.$}
\end{algorithmic}
\end{algorithm}

To solve $P(X),$ we use Algorithm \ref{LP-algorithm}, which is a modification of von Neumann's algorithm for linear programming communicated in \cite{vonneumann}.  As input, it takes $X$ and a {\em state} $S$ defined as follows:
\begin{definition}
A state $S$ is an empty state (denoted as $\emptyset$) or a pair $(h_S, d_S),$ where $h_S\in\mathcal{H}$ and $d_S\in\mathbb{R},$ defining a halfspace which we denote by $H_S.$ For $S = \emptyset,$ we set $H_{\emptyset} = \mathcal{H}.$ That is, $H_{\emptyset}$ is the entire space.
\end{definition}

A state $S\ne\emptyset$ is compatible with $X$ if $(h_S, d_S)$ is a convex combination of a finite subset of $\{y(x)(x, 1) : x\in X\}.$ Given a state $S$ compatible with $X,$ the LP algorithm returns another state $S^\prime$ compatible with $X.$ The final state $S^\prime$ reached by the LP algorithm with the input $(S, X, \gamma)$ will be denoted by $\mbox{LP}(S, X).$ Here we omit $\gamma$ because it is always the same.

The algorithm works in the product space $\mathcal{H}\times\mathbb{R}$ where the inner product is defined as $\inprod{(h_1, d_1)}{(h_2, d_2)} = \inprod{h_1}{h_2}_{\mathcal{H}} + d_1d_2.$ Here, $\inprod{\cdot}{\cdot}_{\mathcal{H}}$ refers to the inner product in the original space $\mathcal{H}.$ The norm in the product space is induced by its inner product. 

In the while-loop of the LP algorithm there is a step which we will call a progress-contributing step. This name is motivated by the fact that, whenever the LP algorithm reaches such a step, it updates $(h^\prime, d^\prime)$ so that $\|(h^\prime, d^\prime)\|^{-2}$ increases by a guaranteed value. Since $\|(h^\prime, d^\prime)\|^{-2}$ is upper bounded by some fixed value depending on $\gamma,$ this allows us to estimate the number of iterations of the LP algorithm (i.e., the number of iterations of its while-loop); here, we adapt the analysis of the progress of the algorithm proposed in \cite{Chubanov2015} for the finite-dimensional case. The respective statements are formulated in the following lemma:



\begin{figure}
	\begin{center}
		\setlength{\unitlength}{1cm}
		\includegraphics[width=0.4\textwidth]{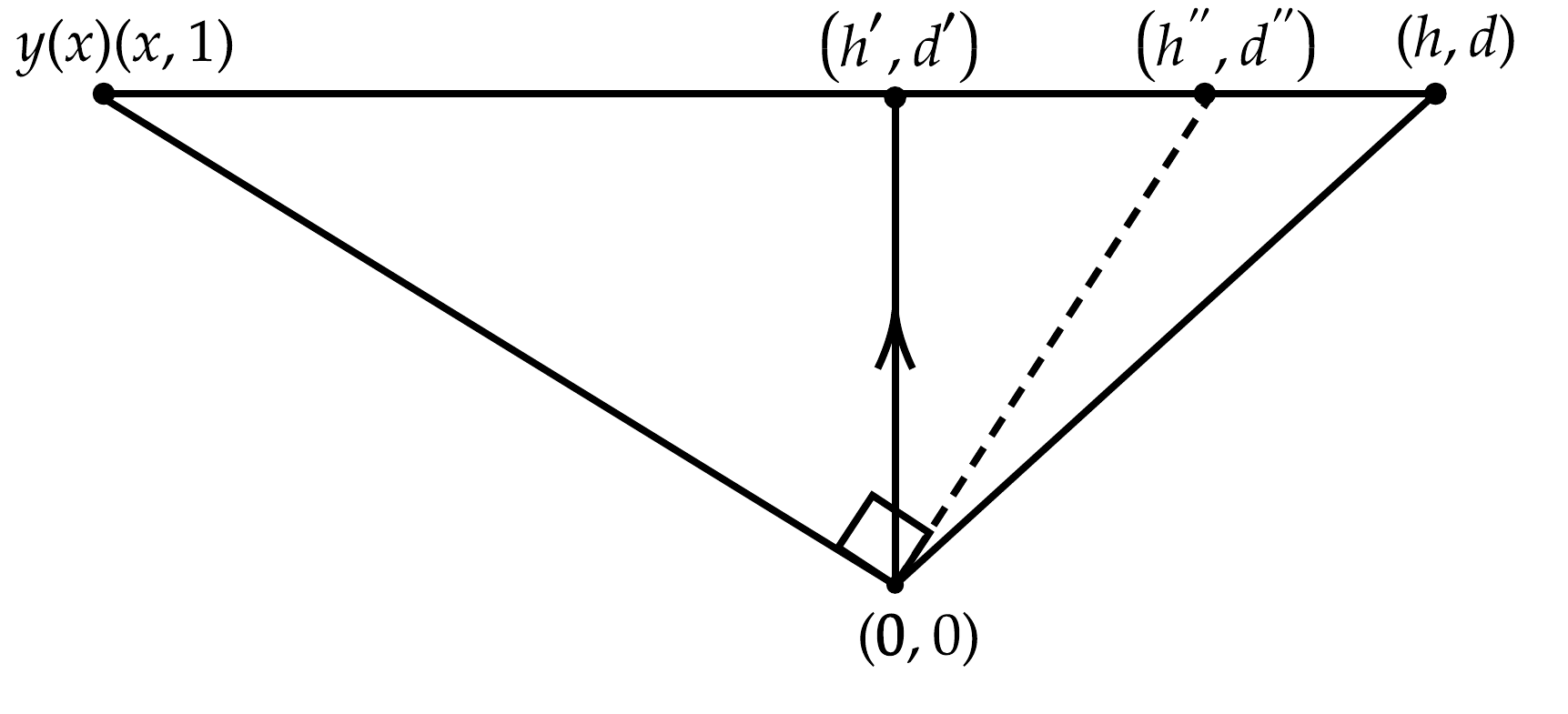}
	\end{center}
	\caption{Iteration of the LP algorithm. 
	}\label{iteration-von-Neumann-picture}
\end{figure}


\begin{lemma}\label{von-neumann-lemma}
The following statements are true with respect to the LP algorithm:
\begin{itemize}
\item[(a)] At a progress-contributing step, the inverse squared length of $(h, d)$ increases by at least that of $y(x)(x, 1):$ 
\begin{equation}\label{increment} 
\|(h^\prime, d^\prime)\|^{-2} \ge \|(h, d)\|^{-2} + \|y(x)(x, 1)\|^{-2}.
\end{equation}
\item[(b)] The following property is invariant during the course of Algorithm \ref{LP-algorithm}, provided that  $X$ is linearly $\gamma$-separable:
\begin{equation}\label{invariant}
\|(h^\prime, d^\prime)\| \ge \gamma/2
\end{equation} 
\item[(c)] If Algorithm \ref{LP-algorithm} reports that $X$ is not linearly $\gamma$-separable, this decision is correct.
\item[(d)] The number of iterations performed by the LP algorithm is bounded by $O(\gamma^{-2}).$
\item[(e)] If $X$ is linearly $\gamma$-separable, then the algorithm returns a state $S^\prime$ such that the associated halfpsace $H_{S^\prime}$ correctly separates $X^-$ and $X^+.$
\end{itemize}
\end{lemma}
{\bf Proof.} (a) This follows from the reciprocal Pythagorean theorem, which holds also for Hilbert spaces. Since $x$ corresponds to a violated inequality of $P(X),$ we have \[\inprod{(h, d)}{y(x)(x, 1)} = y(x)(\inprod{x}{h} + d) \le 0.\] Therefore, the triangle with vertices $(h, d), ({\bf 0}, 0),$ and $y(x)(x, 1)$ contains a right triangle with the right angle at $({\bf 0}, 0)$ and whose the other two vertices are $y(x)(x, 1)$ (one of the vertices of the previous triangle) and some $(h^{\prime\prime}, d^{\prime\prime})\in [(h,d), y(x)(x, 1)],$ i.e., a point on the side opposite to $({\bf 0}, 0),$ of the previous triangle. Both triangles share the same height, which is the segment joining $({\bf 0}, 0)$ and its orthogonal projection $(h^\prime, d^\prime)$ on $[(h, d), y(x)(x, 1)].$ See Figure \ref{iteration-von-Neumann-picture} for an illustration.

(b) Since the projection onto a line segment is a convex combination of the endpoints of the segment, at each iteration $(h^\prime, d^\prime)$ belongs to the convex hull of $\{y(x)(x, 1) : x\in X\},$ which means that there exists $c:X\longrightarrow [0, 1]$ such that
\[
(h^\prime, d^\prime) = \left(\sum_{x\in X} c(x)y(x)x, \sum_{x\in X} c(x)y(x)\right), \sum_{x\in X} c(x) = 1.
\]
Since $X$ is linearly $\gamma$-separable (by the assumption of (b)) and is contained in the unit ball centered at ${\bf 0},$ there exists a solution $(h^*, d^*)$ of $P(X)$ with $\|h^*\| = 1$ and $|d^*| \le 1$ such that $y(x)(\inprod{x}{h^*} + d^*) \ge \gamma$ for all $x\in X.$ 
It follows that
\[
\gamma = \sum_{x\in X} c(x)\gamma \le \sum_{x\in X} c(x)y(x)(\inprod{x}{h^*} + d^*) 
\]
\[=
\inprod{(h^\prime, d^\prime)}{(h^*, d^*)}\le \|(h^\prime, d^\prime)\|\|(h^*, d^*)\|.
\]
Since $\|(h^*, d^*)\|\le 2$ and $(h^*, d^*)$ is feasible, and $\|x\| \le 1$ for all $x\in\mathcal{X},$ this implies (\ref{invariant}).

(c) follows from (b) and (d) follows from (a) and (b).

(e) In this case, the algorithm stops only when a feasible solution of $P(X)$ is found.
 \BlackBox

\section{Proper polyhedral separation in Hilbert spaces}

From the theory of support vector machines, it follows that one can always find a suitable kernel such that the underlying data are linearly separable when represented in the respective RKHS. For this purpose, we can take, e.g., a radial-basis-function kernel. Such kernels are called universal kernels. So the case of linear separable classes is general enough. However, in many situations we would prefer a given kernel or it may be computationally infeasible to find one leading to linear separation, in a given class of kernels. This motivates us to study the case when $\mathcal{X}^-$ and $\mathcal{X}^+$ can be separated by a polyhedron. Another important aspect of polyhedral separation is the mentioned realizability in a suitable RKHS by an explicit choice of a kernel being a function of a given kernel. 

Given a set $X\subset\mathcal{X},$ the major difficulty of the polyhedral classification problem is related to a correct representation of $X^-$ as a union of $t$ sets $X^-_i,$ $i\in[t],$ in the sense that there exists a $\gamma$-separating polyhedron defined by some halfspaces $H_i,$ $i\in [t],$ such that $X^-_i\cap H_i = \emptyset.$ If we knew such a representation, we would solve the problem by simply applying Algorithm \ref{LP-algorithm} $t$ times. Unfortunately, the NP-hardness of proper learning even in the case $t = 2$ suggests that no polynomial algorithm exists for finding a correct representation of this type, unless $P = NP.$ 

At this stage we need to fix a hypothetical "ground-truth" polyhedron which we will need for theoretical purposes. That is, 
let us fix a $\gamma$-separating $t$-polyhedron $F^*$ which correctly classifies the entire instance space $\mathcal{X}.$ Polyhedron $F^*$ contains $\mathcal{X}^+$ and is not intersected with $\mathcal{X}^-.$ Let $F^*$ be defined by halfspaces $H_i,$ $i\in[t].$ Then $\mathcal{X}^+\subset H_i$ for all $i\in[t].$ On the other hand, $\mathcal{X}^-$ can be represented as a union of some of its subsets $\mathcal{X}^-_i,$ $i\in[t],$ such that $H_i\cap\mathcal{X}^-_i = \emptyset$ for each $i\in[t].$ Each set $\mathcal{X}^+\cup \mathcal{X}^-_i$ is linearly $\gamma$-separated by the respective halfspace $H_i.$  

From the above construction, only the union (where the subsets may overlap with each other)
\[
\mathcal{X}^- = \mathcal{X}_1\cup\ldots\cup\mathcal{X}_t
\] 
is assumed to be fixed. We will use this notation throughout this section. (The notation for halfspaces $H_i$ is not fixed and may further have a different meaning depending on the context.) 

There can be many ground-truth polyhedra with the above properties; we only assume that at least one exists. It should be stressed again that we need this hypothetical representation only for the theoretical analysis of the algorithm.

\begin{definition}\label{i-correct-definition}
A set $A$ is called $i$-correct if $A^-\subset\mathcal{X}^-_i.$ (Note that $A$ can be $i$-correct for different $i.$) 
\end{definition}

Further, by a $t$-partition of a set $\tilde{X}$ we mean a tuple $(\tilde{X}_1,\ldots, \tilde{X}_t)$ of its subsets whose union is $\tilde{X}.$ In this definition, a set in the tuple is allowed to be empty. 

Let $\mathcal{F}_i,$ $i\in[t],$ be some families of sets in $\mathcal{H}.$ Consider the following condition:
\begin{condition}\label{partition-condition}
There exists a $t$-partition $(\tilde{X}_1,\ldots, \tilde{X}_t)$ of  $\tilde{X}$ such that for each  $i\in[t]$ with $\tilde{X}_i\ne\emptyset$ there exists $H_i\in\mathcal{F}_i$ such that $H_i\cap \tilde{X}_i = \emptyset.$
\end{condition}

Relatively to polyhedral separation, Condition \ref{partition-condition} has the following implication, when each element of $\mathcal{F}_i$ is a halfspace or the entire space $\mathcal{H}$: If Condition \ref{partition-condition} is satisfied, then there exist sets $H_i\in\mathcal{F}_i,$ $i\in[t],$ whose intersection does not contain any of the sets $\tilde{X}_i.$ In the context of polyhedral separation of a given sample $X,$ we will be interested in the case where each of the sets of the families $\mathcal{F}_i$ contains $X^+$ and $\tilde{X}$ is the set $X^-$ of negative instances. Then, if Condition \ref{partition-condition} is satisfied, the respective intersection $\cap_{i\in[t]}H_i$ yields a $t^\prime$-polyhedron, with $t^\prime\le t,$ consistently partitioning $X.$ (We have $t^\prime \le t$ because $H_i$ can be $\mathcal{H}$ according to our construction).

At each of its iterations, our separation algorithm (Algorithm \ref{separation-algorithm}) tries to construct a consistent $t^\prime$-polyhedron with $t^\prime \le t.$ In the case of success, it returns the polyhedron found in this way, otherwise Condition \ref{partition-condition} is not satisfied for some suitably chosen families $\mathcal{F}_i$ (implicit in the algorithm, but used later for the theoretical analysis) and the algorithm continues. The logic of the algorithm is motivated by the following lemma which holds when Condition \ref{partition-condition} is not satisfied:

\begin{lemma}\label{combinatorial-lemma} 
If Condition \ref{partition-condition} is not satisfied for $\tilde{X},$ then for each $t$-partition $(\tilde{X}_1,\ldots, \tilde{X}_t)$ of  $\tilde{X}$ there exists $j\in[t]$ with $\tilde{X}_j\ne\emptyset$ such that 
\begin{equation}\label{progress-condition}
\forall H\in\mathcal{F}_j: H\cap \tilde{X}_j \neq\emptyset.
\end{equation}
\end{lemma}
{\bf Proof.} 
The lemma is obtained directly by negating Condition \ref{partition-condition}.
\BlackBox



\begin{algorithm}[tb]
	\caption{\bf Separation algorithm}
	\label{separation-algorithm}
\begin{algorithmic}
\STATE{{\bf Input: } $X,$ $t$ and $\gamma.$}
\STATE{{\bf Output:} A consistent $t$-polyhedron if there exists a $\gamma$-separating $t$-polyhedron.}
\STATE{$\mathcal{A}:= \{X^+\}$}
\STATE{$S(X^+, 0) := \emptyset$ (initial state at the root of the search tree)}
\WHILE{no polyhedron found}
\STATE{$k := k + 1$}
\STATE{Find out if there are $A_1,\ldots, A_{t^\prime}\in\mathcal{A}$ with $t^\prime\le t$ such that $F = \cap_{p\in[t^\prime]} H_{S(A_p, k - 1)}$ correctly classifies $X.$}
\IF{there is such $F$}
\STATE{Return $F$}
\ENDIF
\STATE{(Branching step)}
\STATE{Set $S(A, k):= S(A, k -1)$ for all $A\in\mathcal{A}.$}
\FOR{ all $(A, x)\in\mathcal{A}\times X^-$}
\IF{$x\in H_{S(A, k - 1)}$}
\STATE{$S(A\cup\{x\}, k) := \mbox{LP}(S(A, k - 1), A\cup\{x\})$}
\IF{ the LP algorithm does not report that $A\cup \{x\}$ is not linearly $\gamma$-separable}
\STATE{Add $A\cup \{x\}$ to $\mathcal{A}.$}
\ENDIF
\ENDIF
\ENDFOR
\ENDWHILE
\end{algorithmic}
\end{algorithm}

For the analysis of Algorithm \ref{separation-algorithm}, we introduce the notion of the {\em search tree}. Each level of the search tree corresponds to an iteration of the main loop of the algorithm, where a certain set family $\mathcal{A}$ is considered. The family $\mathcal{A}$ at the end of the $k$th iteration corresponds to the set of nodes forming the $k$th level of the search tree. Each node of the search tree is encoded by a pair $(A, k)$ where $A$ is contained in $\mathcal{A}$ at the end of iteration $k,$ i.e., $k$ is the level, of the search tree, containing $(A, k).$ The nodes of the neighboring levels can be connected by arcs. There is an arc from $(A, k - 1)$ to $(A^\prime, k)$ if and only if $A = A^\prime$ or $A^\prime = A + \{x\}$ for some $x.$ 

When the algorithm decides whether a new node of the form $(A\cup\{x\}, k)$ is to be added to the search tree, based on a node $(A, k - 1)$ at the previous level, it solves the respective problem $P(A\cup\{x\}).$ Then, $(A\cup\{x\}, k)$ appears in the search tree if the LP algorithm finds a halfspace consistently partitioning $A\cup\{x\}$ (i.e., if the LP algorithm does not report that $A\cup\{x\}$ is not linearly $\gamma$-separable).
The algorithm stores the states $S(A, k)$ reached by the LP algorithm at the respective nodes of the search tree. Informally, our algorithm works as follows:
\begin{itemize}
\item Given a sample $X,$ the algorithm tries to construct a separating polyhedron defined by no more than $t$ inequalities. To "assemble" such a polyhedron, it tries to find a suitable combination of halfspaces associated with the states stored at the current level. 
\item At each iteration, it expands the search tree by adding a new level. The nodes of the previous level stay in the new one. At the branching step, for each node $(A, k - 1),$ the algorithm generates a new node $(A\cup\{x\}, k),$ provided that $x$ is not separated by the halfspace $H_{S(A, k -1)}$ associated with the parent node $(A, k - 1).$ The LP algorithm is then applied to the problem $P(A\cup\{x\})$ starting with the state $S(A, k - 1)$ that has been previously reached for the parent node. Since $H_{S(A, k - 1)}$ does not consistently separate $x,$ it follows that the LP algorithm makes at least one progress-contributing step. We will use this property in the analysis of the running time of the algorithm.

\item The search tree has the property that the state associated with a node $(A, k)$ is compatible with the states associated with the other nodes of the subtree rooted at $(A, k).$ This makes it possible to trace the progress of the LP algorithm along each path of the search tree. For this purpose, we introduce the notion of a progress-contributing arc, which is one corresponding to a transition via the LP algorithm from a state to another one with at least one progress-contributing step of the LP algorithm. 
\end{itemize}

\begin{figure}
	\begin{center}
		\setlength{\unitlength}{1cm}
		\includegraphics[width=0.5\textwidth]{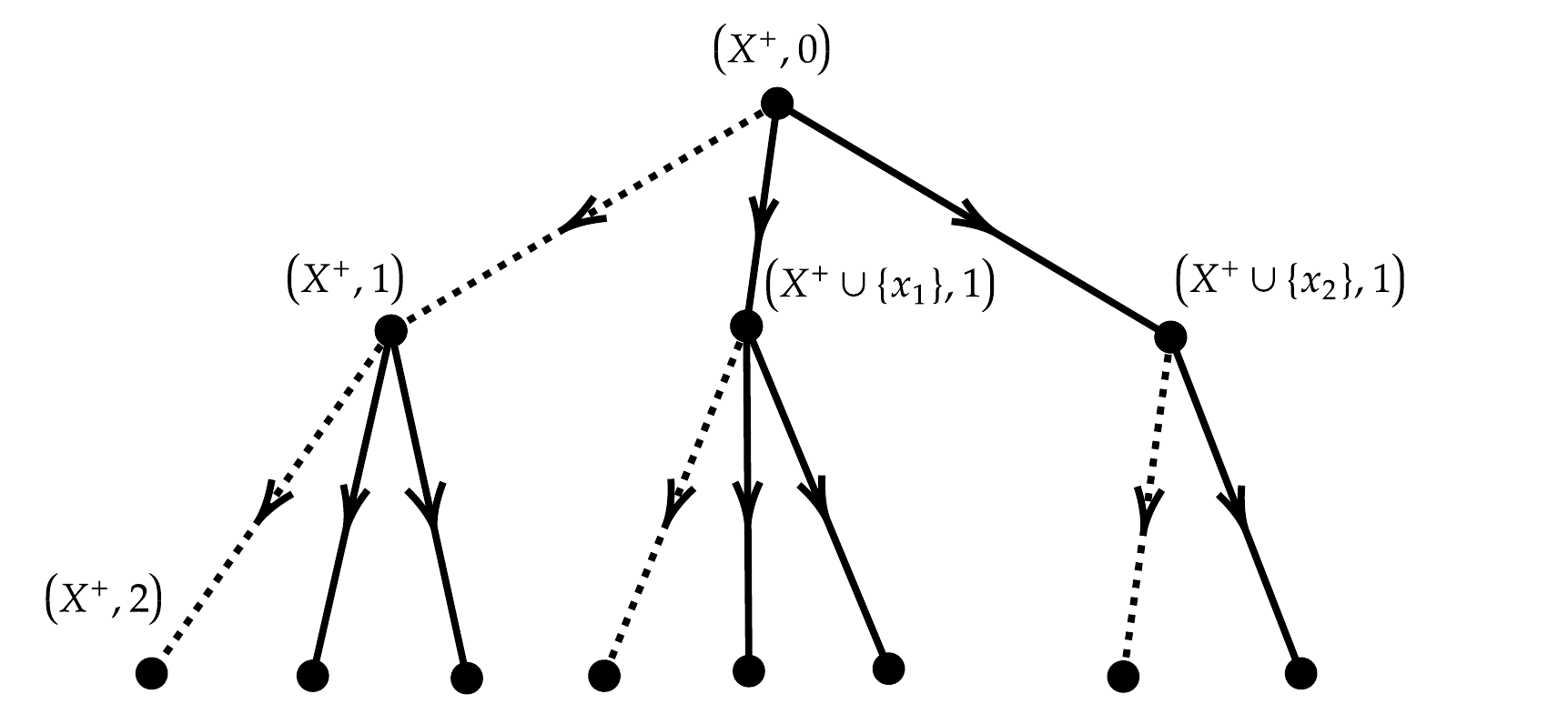}
	\end{center}
	\caption{Search tree: solid arcs are progress-contributing; for each parent node, if more than one arc during the branching step, then one of them is not progress-contributing and the others should be progress-contributing. 
	}\label{search-tree-picture}
\end{figure}

An example of a search tree constructed by Algorithm \ref{separation-algorithm} is illustrated in Figure \ref{search-tree-picture}.

Recall that each nonempty state $S$ consists of an associated pair $(h_S, d_S)$ defining a halfspace $H_S.$ When reaching another state $S^\prime\ne S$ from $S,$ the LP algorithm makes at least one progress-contributing step that increases $\|(h_S, d_S)\|^{-2}$ by a certain guaranteed value ($1/4$) following from inequality (\ref{increment}) of Lemma \ref{von-neumann-lemma}. So it is natural to introduce the following potential function $\pi$ defined at each node of the search tree:
\[
\pi(A, k) := \left\{
\begin{array}{ll} 
\|(h_{S(A, k)}, d_{S(A, k)})\|^{-2}, & \mbox{if } S(A, k)\ne\emptyset,\\
0, & \mbox{otherwise.}
\end{array}
\right.
\]
We call an arc $((A, k - 1), (A^\prime, k))$ progress-contributing if 
\[
\pi(A^\prime, k) \ge \pi(A, k - 1) + 1/4.
\]
This inequality takes place if and only if the LP algorithm makes at least one progress-contributing step when starting with $S(A, k - 1)$ to solve the problem $P(A^\prime).$ If $A^\prime = A,$ then $S(A^\prime, k) := S(A, k - 1).$ Therefore,
for each arc $((A, k - 1), (A^\prime, k))$ of the search tree, 
$
\pi(A^\prime, k) \ge \pi(A, k - 1).
$
That is, $\pi$ is non-decreasing along each path of the search tree.

On the other hand, the LP algorithm implies that 
\[
\pi(A, k) \le 4\gamma^{-2}.
\]
Therefore, the number of progress-contributing arcs on each path of the search tree is bounded by $O(\gamma^{-2}).$

%
%

\begin{lemma}\label{tree-width-lemma}
At the end of iteration $k,$ $|\mathcal{A}|$ is bounded by 
\begin{equation}\label{tree-width}
((m + 1)k)^{O(\gamma^{-2})},
\end{equation}
where $m = |X|.$
\end{lemma}
{\bf Proof.} Since each path of the search tree contains no more than $O(\gamma^{-2})$ progress-contributing arcs, it follows that each path from the root to the $k$th level of the tree can be uniquely encoded by a $k$-vector whose components take integer values from $0$ to $m$ and whose number of nonzero components is bounded by $O(\gamma^{-2}).$  The (\ref{tree-width}) is an upper bound on the total number of such $k$-vectors. At the same time, this is an upper bound on the number of nodes at the $k$th level of the search tree, i.e., on $|\mathcal{A}|$ at the end of the $k$th iteration. \BlackBox

Now we come back to the notion of $i$-correctness (Definition \ref{i-correct-definition}) and extend it to the notion of $i$-correctness of a path: Further, we say that a path in the search tree is $i$-correct if $A$ is $i$-correct for each node $(A, k)$ on this path. 

\begin{lemma}\label{i-correct-path-lemma}
If $(A, k)$ is a node of the search tree and $A$ is $i$-correct, then the path connecting the root and $(A, k)$ is $i$-correct.
\end{lemma}
{\bf Proof.} The path from the root has the form \[(A_0, 0) = (X^+, 0), (A_1, 1),\ldots, (A_k, k) = (A, k).\] Observe that $A_l\subseteq A_{l +1},$ $l = 0,\ldots, k - 1.$ This means that if some $A_l$ is not $i$-correct then $A_k$ is not $i$-correct because $A_l$ is contained in $A_k,$ which contradicts the assumption that $A$ is $i$-correct. \BlackBox


\begin{theorem}\label{separation-theorem} If there is a consistent $\gamma$-separating $t$-polyhedron, then 
the running time of Algorithm \ref{separation-algorithm} is bounded by 
\begin{equation}\label{proper-complexity}
m^{O(t\gamma^{-2}\lceil\log_m(\gamma^{-1})\rceil)}.
\end{equation}
\end{theorem}
{\bf Proof.} Consider iteration $k.$ Let $\tilde{X} = X^-.$ Consider a $t$-partition $(\tilde{X}_1,\ldots, \tilde{X}_t)$ of  $\tilde{X}$ such that each nonempty $\tilde{X}_i$ is $i$-correct. For each $i\in[t]$ with $\tilde{X}_i\ne\emptyset,$ let 
$\mathcal{F}_i$ be the family of all halfspaces associated with states $S(A, k - 1)$ corresponding to $i$-correct sets $A$ in $\mathcal{A}.$ For each $i$ with $\tilde{X}_i = \emptyset,$ let $\mathcal{F}_i =\{\mathcal{H}\}.$ 

Let $\mathcal{A}_i$ denote the family of all $i$-correct sets in $\mathcal{A}.$ Note that the family $\mathcal{A}_i$ is not empty for each $i\in[t]$ because it at least contains $X^+.$

Assume that the algorithm does not find a consistent $t^\prime$-polyhedron with $t^\prime \le t$ at the current iteration. Then no consistent $t$-polyhedron can be assembled from the families $\mathcal{F}_i,$ $i\in[t].$ It follows that Condition \ref{partition-condition} is not satisfied. Then Lemma \ref{combinatorial-lemma} implies that there exists $j\in[t]$ such that $\tilde{X}_j\cap H_{S(A, k - 1)}\ne\emptyset$ for all $j$-correct sets $A\in\mathcal{A}.$ This implies that, for each $A\in\mathcal{A}_j,$ the current iteration $k$ should construct a progress-contributing arc connecting $(A, k - 1)$ with $(A^\prime, k)$ where $A^\prime = A\cup\{x\},$ $x\in\tilde{X}_j.$ That is, $A^\prime$ is also $j$-correct. Denote this $j$ by $j_k,$ where $k$ is the number of the current iteration.

Now let us consider the sequence of indices $j_1, j_2, \ldots$ over all iterations of the algorithm. Consider the subsequence of indices equal to some $r\in[t]:$ $j_{k_1} = r, j_{k_2} = r,\ldots.$
Here, $k_1,k_2,\ldots$ are the respective levels of the search tree. For each $k_s,$ the previous level $k_s - 1$ in the search tree contains a node $(A, k_s - 1)$ where $A$ is $r$-correct. Lemma \ref{i-correct-path-lemma} implies that the path from the root to $(A, k_s - 1)$ is $r$-correct. Our construction implies that each $r$-correct path from the root to level $k_s -1$ is connected with level $k_s$ by a set of arcs where at least one arc is progress-contributing and has the form $((A, k_s - 1), (A^\prime, k_s))$ where $A^\prime$ is $r$-correct, for each $s.$ It follows that there exists an $r$-correct path, over all levels of the search tree, which contains at least as many progress-contributing arcs as the number of elements of the subsequence.

Assume that the number of iterations is infinite or exceeds $4t\gamma^{-2}.$ Then, for some $l,$ the subsequence of indices equal to $l$ contains more than $4\gamma^{-2}$ elements. This implies a contradiction because in this case the search tree should contain an $l$-correct path with more than $4\gamma^{-2}$ progress-contributing arcs. Therefore, the number of iterations is not greater than $4t\gamma^{-2}.$ When the algorithm stops, it returns a consistent $t^\prime$-polyhedron with $t^\prime\le t$ because this is the only condition when it can stop.

The number of elementary operations needed for the LP algorithm whenever it is called from Algorithm \ref{separation-algorithm} is bounded by $O(m\gamma^{-2})$ where the sample size $m = |X|$ appears because, at each of its iterations, the LP algorithm needs to find a violated constraint of a problem of the form $P(A\cup\{x\}),$ where $(A\cup\{x\})\subseteq X.$ At each iteration, the time needed to construct $F$ or make sure that none exists is bounded by $O(|\mathcal{A}|^t).$ Since the number of iterations of Algorithm \ref{separation-algorithm} is bounded by $O(t\gamma^{-2}),$ Lemma \ref{tree-width-lemma} implies that $|\mathcal{A}|$ is bounded by (\ref{tree-width}) with $k$ replaced by $t\gamma^{-2}.$  Replacing $\gamma^{-2}$ by $m^{\log_m(\gamma^{-2})}$ and rearranging the terms, we get the complexity bound (\ref{proper-complexity}).
\BlackBox

Now we formulate another algorithm based on similar ideas. In contrast to Algorithm \ref{separation-algorithm}, it constructs $F$ as the intersection of all halfspaces available at the current iteration. If $F$ is not consistent with the given sample $X,$ then it chooses an arbitrary element $x$ of $X$ to which $F$ assigns a false label. Since all the halfspaces contain $X^+,$ it follows that $x\in X^-.$ Since $x\in F,$ it follows that $x\in H_{S(A, k - 1)}$ for all $A\in\mathcal{A}$ at the beginning of the current iteration. This observation allows us to prove that the previous bound on the number of iterations of Algorithm \ref{separation-algorithm} is also valid for Algorithm \ref{improper-separation-algorithm}.

\begin{algorithm}[tb]
	\caption{\bf Improper-separation algorithm}
	\label{improper-separation-algorithm}
\begin{algorithmic}
\STATE{{\bf Input: } $X$ and $\gamma.$}
\STATE{{\bf Output:} A polyhedron consistent with $X$ if there exists a $\gamma$-separating polyhedron.}
\STATE{$\mathcal{A}:= \{X^+\}$}
\STATE{$S(X^+, 0) := \emptyset$ (initial state at the root of the search tree)}
\WHILE{no polyhedron found}
\STATE{$k := k + 1$}
\STATE{$F := \cap_{A\in\mathcal{A}} H_{S(A, k - 1)}$}
\IF{$F$ correctly separates $X^-$ and $X^+$}
\STATE{Return $F$}
\ENDIF
\STATE{Pick an arbitrary $x\in X$ incorrectly classified by $F.$ ($x\in X^-$ because $X^+\subset F.$}
\STATE{(Branching step)}
\STATE{Set $S(A, k):= S(A, k -1)$ for all $A\in\mathcal{A}.$}
\FOR{ all $A\in\mathcal{A}$}
\STATE{$S(A\cup\{x\}, k) := \mbox{LP}(S(A, k - 1), A\cup\{x\})$}
\IF{ the LP algorithm does not report that $A\cup \{x\}$ is not linearly $\gamma$-separable}
\STATE{Add $A\cup \{x\}$ to $\mathcal{A}.$}
\ENDIF
\ENDFOR
\ENDWHILE
\end{algorithmic}
\end{algorithm}

\begin{theorem} If there exists a $\gamma$-separating $t$-polyhedron correctly classifying the instance space, then Algorithm \ref{improper-separation-algorithm} finds a consistent polyhedron in time
\begin{equation}\label{improper-complexity-bound}
O\left(m\cdot(2t\gamma^{-2})^{4\gamma^{-2}}\right)
\end{equation}
The number of halfspaces defining the polyhedron is bounded by 
\begin{equation}\label{number-of-halfspaces}
O((8t\gamma^{-2})^{4\gamma^{-2}}).
\end{equation}
\end{theorem}
{\bf Proof.} The number of iterations is bounded by the same value as that for Algorithm \ref{separation-algorithm}. This can be proved by the same argument as that we have used in the proof of Theorem \ref{separation-theorem}. The only difference is how polyhedron $F$ is constructed and what conclusions are made if no polyhedron $F$ is found; in Algorithm \ref{improper-separation-algorithm}, we simply check if the polyhedron $F,$ built as the intersection of halfspaces associated with all states reached at the previous level, gives us a consistent polyhedron. If it doesn't, then each arc $((A, k - 1), (A\cup\{x\}, k))$ is progress-contributing because $x\in X^-$ is contained in each of the halfspaces $H_{S(A, k - 1)},$ which means that the LP algorithm makes at least one progress-contributing step during the transition from $S(A, k - 1)$ to $S(A\cup\{x\}, k).$ Now we can derive an upper bound $4t\gamma^{-2}$ on the number of iterations in the way identical to that used in the proof of Theorem \ref{separation-theorem}. To estimate $|\mathcal{A}|$ at the end of each iteration, we use (\ref{tree-width}) where $k$ should be replaced by  $4t\gamma^{-2}$ and $m + 1$ by $2$ because the search tree constructed by Algorithm \ref{improper-separation-algorithm} is binary, which means that the paths can be encoded by binary vectors. \BlackBox

{\em Remark.} If none consistent $\gamma$-separating $t$-polyhedron exists, then the number of iterations of both algorithms will be infinite. If such a polyhedron exists, the number of iterations of both algorithms is bounded by $4t\gamma^{-2}.$ So we can add a step verifying whether $k\le 4t\gamma^{-2}.$ If this is not the case, then no consistent $\gamma$-separating $t$-polyhedron exists.
\section{PAC polyhedral classification}
If $\mathcal{H}$ is finite-dimensional, the VC dimension of the $t$-polyhedron returned by Algorithm \ref{separation-algorithm} is bounded by $O(dt\log t).$ To obtain the VC dimension of the polyhedron returned by Algorithm \ref{improper-separation-algorithm}, the $t$ in the above estimate should be replaced by the respective bound (\ref{number-of-halfspaces}). Thus, we come to the following corollary:
\begin{corollary} Let $\mathcal{H}$ be a space of dimension $d < \infty.$ Then the following statements are true:
\begin{itemize}
\item Algorithm \ref{separation-algorithm} constructs a PAC polyhedron in time $m^{O(t\gamma^{-2})}$ from a concept class whose VC dimension $D$ is bounded by $O(dt\log t).$
\item Algorithm \ref{improper-separation-algorithm} constructs a PAC polyhedron in time (\ref{improper-complexity-bound}) from a concept class with VC dimension $D$ bounded by $\vcdimension.$ 
\end{itemize}
\end{corollary} 

The above PAC learning is then possible with a sample complexity of $O(\varepsilon^{-1}(D + \log(\delta^{-1})),$ where $D$ is the respective VC dimension.

For the infinite-dimensional case, we prove the following:
\begin{corollary}
Let $\mathcal{H}$ be an RKHS with a continuous kernel $K.$ Let $\mathcal{X}$ be the image of a compact set $\Omega$ in $\mathbb{R}^s$ under the feature map $\varphi$ of the RKHS. Both Algorithm \ref{separation-algorithm} and Algorithm \ref{improper-separation-algorithm} can be implemented so as to produce PAC polyhedra under the same assumptions as before. The sample complexity is polyhomially bounded in the dimension $s$ of the space containing $\Omega,$ $\gamma^{-1},$ and $t.$
\end{corollary}
{\bf Proof.} The main idea is to use the well-known discretization trick which consists in replacing $\Omega$ by a finite set $\Omega^\#$ with a sufficiently small step size. Then the algorithms are modified by replacing elements $x$ of $\mathcal{H}$ occurring in the course of the algorithms by their approximations from the image $\varphi(\Omega^\#).$ Of course, $\varphi$ is not computed explicitly because we only need $K$ to compute inner products. For more details, see the appendix. \BlackBox

\section{Summary}

We propose an algorithm for PAC learning in Hilbert spaces, such as
RKHS's. The algorithm learns polyhedral concepts, which makes it universally applicable because a suitable kernel can always be explicitly chosen so that polyhedral concepts become realizable in the respective RKHS.

\bibliographystyle{plainnat}

\section{Appendix}

\subsection{Universality of polyhedral classification}

Let us first consider a special case where $\mathcal{X}$ is a subset of a Boolean cube $\{0, 1\}^n.$ Any partition of $\mathcal{X}$ into $\mathcal{X}^-$ and $\mathcal{X}^+$ can be correctly classified by a suitable conjunctive normal form (CNF), which can be expressed in the form of a finite system of linear inequalities whose solution set contains $\mathcal{X}^+$ and does not contain any instance of $\mathcal{X}^-.$ Let $Ax \le b$ be such a system with $t$ inequalities, where $A$ is a coefficient matrix and $b\in\mathbb{R}^t.$ Then $\mathcal{X}^-$ can be represented as $\mathcal{X}^- = \mathcal{X}_1\cup\cdots\cup\mathcal{X}_t,$ where $\mathcal{X}_i$ consists of points in $\mathcal{X}^-$ violating the $i$-th constraint. Let $2\gamma_i$ be the distance between the convex hull of $\mathcal{X}^-_i$ and that of $\mathcal{X}^+.$ Let $\gamma = \min_i \gamma_i.$ Then $\mathcal{X}^-$ can be $\gamma$-separated from $\mathcal{X}^+$ by the $t$-polyhedron defined by $Ax \le b.$

Thus, if we deal with an instance space where instances are encoded by binary vectors of length not greater than some $d,$ then the above binary case implies that such data can be correctly classified by a suitable polyhedron in the space of binary encodings.

Now assume that we have to consistently separate some subset $\Omega$ in the unit ball centered at the origin in some Euclidean space of dimension $d.$ We additionally assume that the opposite classes $\Omega^-$ and $\Omega^+$ are distinguishable in the sense that there is $\rho > 0$ such that $\|\omega_0 - \omega_1\| \ge \rho$ for all $\omega_0\in{\Omega}^-$ and $\omega_1\in{\Omega}^+.$ Let $\mathcal{X}$ consist of vectors $(\omega, \sqrt{1 - \|\omega\|^2}),$ $\omega\in\Omega.$ So $\mathcal{X}$ is the image of $\Omega$ on a sphere, under the respective mapping, in space $\mathbb{R}^{d + 1}.$ Set $\mathcal{X}^-$ can be covered by a finite number of smaller balls $B_i$ in $\mathbb{R}^{d + 1}$ not intersecting $\mathcal{X}^+.$ Denote the number of balls $B_i$ by $t.$ The intersection of each ball $B_i$ with the sphere can be linearly $\gamma_i$-separated from $\mathcal{X}^+,$ for some $\gamma_i > 0.$ Let $\gamma$ be the minimum of all $\gamma_i,$ $i\in[t].$ Then $\mathcal{X}^+$ is contained in a $t$-polyhedron which $\gamma$-separates $\mathcal{X}^-$ from $\mathcal{X}^+.$

The above construction is also applicable to Hilbert spaces. Below we give another one, where the whole space is mapped onto a sphere in another Hilbert space.

Let $\tilde{\mathcal{H}}$ be the reproducing kernel Hilbert space of a positive definite kernel $\tilde{K}.$ Now we wish to homeomorphically transform space $\tilde{\mathcal{H}}$ so that its image would belong to a sphere in some other space. For this purpose, we introduce a new kernel $K$ defined as
\[
K(x,x^\prime) = \frac{\tilde{K}(x,x^\prime) + 1}{\sqrt{\tilde{K}(x,x) + 1}\sqrt{\tilde{K}(x^\prime,x^\prime) + 1}}.
\]
The new kernel $K$ is also positive definite. Defining $K$ in this way, we do not destroy the topology of the original RKHS $\tilde{\mathcal{H}}.$ So let $\mathcal{H}$ be the RKHS of $K.$
If $\tilde{\mathcal{X}}$ is an instance space in $\tilde{\mathcal{H}},$ then its image $\mathcal{X}$ in $\mathcal{H}$ lies on the unit sphere centered at the origin because the squared norm of the image of $x\in\tilde{\mathcal{X}}$ is $K(x, x) = 1.$ If $\tilde{\mathcal{X}}$ is an image of a compact set of a Euclidean space and the opposite classes are distinguishable, then in a similar way as before we can come to a conclusion that $\mathcal{X}^-$ and $\mathcal{X}^+$ can be $\gamma$-separated by a polyhedron in $\mathcal{H},$ for some $\gamma > 0.$




\subsection{PAC learning}

The LP algorithm implicitly learns from a finite-dimensional subset of $\mathcal{H}.$ Let $\mathcal{X}$ be linearly $\gamma$-separable. Now apply the LP algorithm beginning with an empty state for a given sample $X.$ Let $S$ be the state reached. Note that this state is compatible with the entire instance space $\mathcal{X}.$ Assume we have a hypothetical separation oracle able to return a violated constraint of the system $P(\mathcal{X})$ comprising all possible instances. Apply the LP algorithm again beginning with $S$ with respect to $\mathcal{X}.$ Beginning with the empty state, the two applications of the LP algorithm reach the final state $S^\prime$ via state $S$ in a total number of $O(\gamma^{-2})$ iterations. Each iteration of the LP algorithm constructs a new $(h, d)$ as a convex combination of the old one and $y(x)(x, 1),$ where $x\in\mathcal{X}.$ It follows that $h_{S^\prime}$ is a convex combination of a finite subset of $\mathcal{X}$ whose cardinality is bounded by $O(\gamma^{-2}).$ In other words, $h_{S^\prime}$ lies in a subspace whose dimension is bounded by $O(\gamma^{-2}).$ Therefore, for any given sample, the LP algorithm learns from a suitable class of linear concepts of a finite VC dimension.  In the case when $\mathcal{X}$ is linearly $\gamma$-separable, this class contains a ground-truth linear concept correctly partitioning the entire instance space $\mathcal{X}.$ 


In contrast to the finite-dimensional case, we cannot directly conclude that our algorithm constructs a polyhedral classifier belonging to a concept class of a finite VC dimension. Therefore, we need some more assumptions. 

Let us assume that $\mathcal{H}$ is an RKHS of a kernel $K$ such that $K:\mathbb{R}^s\times\mathbb{R}^s\map\mathbb{R}$ is continuously differentiable and $\Omega$ is a subset of the cube $[0,1]^s$ in $\mathbb{R}^s.$ In that case $K$ is Lipschitz-continuous over $\Omega.$ 

So let $\varphi$ be the feature map of $\mathcal{H}.$ It is Lipschitz-continuous on $\Omega.$ Then, for some constant $L,$
\[
\|\varphi(\omega) - \varphi(\omega + \Delta\omega)\|\le L\|\Delta\omega\|.
\]

For $\omega,$ let $\omega^\#$ denote the closest element of the lattice formed by the points of the form $\beta z,$ where $z\in\mathbb{Z}^s$ and $\beta$ is a sufficiently small constant which we calculate below.

Observe that \[\|\omega - \omega^\#\| \le \sqrt{s}\beta.\]
Then, choosing \[\beta \le \gamma/(2\sqrt{s}L),\] we get 
\[
\|\varphi(\omega) - \varphi(\omega^\#)\| \le \gamma/2.
\]
Let $B$ be the unit ball centered at the origin and $proj_B$ denote the orthogonal projection on $B.$ Then,
\[
\|\varphi(\omega) - proj_B(\varphi(\omega^\#))\| \le \gamma/2,
\]
because $\varphi(\omega)$ belongs to $B,$ due to our assumption that $\mathcal{X}$ lies in $B,$ and the projection onto $B$ does not increase the distance to $\varphi(\omega).$

It follows that none of the $\gamma/2$-margins of the halfspaces defining a $\gamma$-separating $t$-polyhedron can contain $proj_B(\varphi(\omega^\#)).$ So we can consistently assume that the label of $proj_B(\varphi(\omega^\#))$ is $y(\varphi(\omega)).$ Our discretized instance space is $\mathcal{X}^\# =proj_B( (\varphi(\Omega^\#)),$ where $\Omega^\# = \{\omega^\# : \omega\in\Omega\}.$  The opposite classes of $\mathcal{X}^\#$ can be $\gamma/2$-separated by a $t$-polyhedron.

Let $X$ be a given sample and $X^\#$ denote the set of all $proj_B(\varphi(\omega^\#))$ with $\omega$ ranging over the elements of $\Omega$ encoding the elements of $X.$ Note that we do not need $\varphi$ to be given explicitly. All the computations can be performed only using the inner products (and therefore $K$) and the encodings $\omega$ of the respective elements of $X.$

If $X$ is linearly $\gamma$-separable, then $X^\#$ is $\gamma/2$-separable. If we detect that it is not, then the original sample $X$ is not separable. The idea of the modification of the LP algorithm that we propose below consists in considering only the elements of $X^\#$ in the training process until a separating halfspace is constructed (in this case, we learn from a finite concept class because $\mathcal{X}^\#$ is finite and the coefficients are appropriately rounded) or the length of $(h, d)$ becomes sufficiently small to conclude that $X^\#$ is not linearly $\gamma/2$-separable and therefore $X$ is not $\gamma$-linearly separable. 

The progress-contributing step of the LP algorithm should be modified as follows:
\begin{itemize}
\item Let $\hat{x}\in X$ correspond to a constraint of $P(X)$ violated by $(h, d):$ The $\hat{x}$ is represented by its encoding $\omega\in\Omega.$
\item Let $x = proj_B(\varphi(\omega^\#)).$
\item Let $(\hat{h}, \hat{d})$ be the projection of $({\bf 0}, 0)$ on the line segment $[(h,d), y(x)(x, 1)].$ It is calculated by the formula
\[
(\hat{h}, \hat{d}) = \alpha (h, d) + (1 - \alpha) y(x)(x, 1)
\]
where
\[
\alpha = \frac{\inprod{y(x)(x, 1)}{y(x)(x, 1) - (h, d)}}{\|y(x)(x, 1) - (h, d)\|^2}.
\]
\item If $\|(\hat{h}, \hat{d})\| \le \gamma/4,$ then return that $X$ cannot be linearly $\gamma$-separated because $X^\#$ is not linearly $\gamma/2$-separable in this case.
\item Compute $(h^\prime, d^\prime)$ by rounding $\alpha:$ \[(h^\prime, d^\prime) := \beta\lfloor\alpha/\beta\rfloor (h, d) + (1 - \beta\lfloor\alpha/\beta\rfloor) y(x)(x, 1).\]
\end{itemize}
Note that $x$ does not necessarily violate the constraint associated with $\hat{x}.$ It is sufficient, that the respective progress in the sense of the potential function $\pi$ is preserved.

Let the respective variant of the LP algorithm be called the modified LP algorithm.


%

It remains to determine a value of $\beta$ that would guarantee
\begin{equation}\label{progress-condition}
\|(h^\prime, d^\prime)\|^{-2} \ge \|(h, d)\|^{-2} + const,
\end{equation}
where $const$ is some positive constant.

Note that  $\|(h^\prime, d^\prime)\| \ge \|(\hat{h},\hat{d})\|$ because $(\hat{h}, \hat{d})$ is the orthogonal projection of $({\bf 0}, 0)$ on the respective line segment and $(h^\prime, d^\prime)$ lies on the same segment. 

First, we estimate how small the (nonnegative) difference $\Delta = \|(h^\prime, d^\prime)\|^2 - \|(\hat{h},\hat{d})\|^2$ should be in order to ensure  (\ref{progress-condition}).
We will use the fact that, at the stage when $(h^\prime, d^\prime)$ is computed, we have $\|(\hat{h}, \hat{d})\| \ge\gamma/2:$
\begin{align*}
\|(h^\prime, d^\prime)\|^{-2} & = (\|(\hat{h}, \hat{d})\|^2 + \Delta)^{-1}\\
& = ( 1 + \Delta\|(\hat{h}, \hat{d})\|^{-2})^{-1}\cdot \|(\hat{h}, \hat{d})\|^{-2}\\
& \ge (1 + 4\Delta\gamma^{-2})^{-1}\cdot\|(\hat{h}, \hat{d})\|^{-2}\\
& = \left( 1 - \frac{4\Delta\gamma^{-2}}{1 + 4\Delta\gamma^{-2}}\right)\|(\hat{h}, \hat{d})\|^{-2}\\
&\ge (1 - 4\Delta\gamma^{-2})\|(\hat{h}, \hat{d})\|^{-2}\\
& \ge (1 - 4\Delta\gamma^{-2})(\|(h, d)\|^{-2} + 1/4)\\
& = \|(h, d)\|^{-2} + 1/4 - 4\Delta\gamma^{-2}(\|(h, d)\|^{-2} + 1/4)\\
& = \|(h, d)\|^{-2} + 1/4 - \Delta(4\gamma^{-2}\|(h, d)\|^{-2} + \gamma^{-2})\\
&\ge \|(h, d)\|^{-2} + 1/4 - \Delta(16\gamma^{-2} + \gamma^{-2}).
\end{align*}
Thus, it is sufficient to choose $\beta$ so that $\Delta \le \gamma^2/(8\cdot 17),$ in which case (\ref{progress-condition}) is valid for $const = 1/8.$ Taking into account that $\|(x, 1)\|^2 \le 2,$ we observe that $\Delta\le 4\beta^2.$ Therefore, we can choose $\beta\le \gamma^2/\sqrt{4\cdot 8\cdot 17}.$ Combining with the previous bound, we can set 
\[
\beta = \min\{ \gamma^2/\sqrt{4\cdot 8\cdot 17}, \gamma/(2\sqrt{s}L)\}.
\]

The modified LP algorithm performs no more than $O(\gamma^{-2})$ iterations. The respective class of linear halfspaces is finite. Each of them is defined by a linear combination of no more than $O(\gamma^{-2})$ points of the form $(proj_B(\omega^\#), 1)$ $\omega\in\Omega.$ 

It is now not hard to estimate the number of such combinations taking into account the rounding of coefficients $\alpha$ and the choice of $\beta.$ 

Note that $|\Omega^\#|$ is bounded by $O(\beta^{-s})$ because $\Omega$ is a subset of $[0,1]^s.$ 

There are $O(\beta^{-1})$ variants for the coefficient $\beta\lfloor\alpha/\beta\rfloor$ and $|\Omega^\#|$ possible choices of $x.$ Therefore, independently of the sample in question, the modified LP algorithm chooses linear combinations from 
$(\beta^{-1})^{O(s\gamma^{-2})}$ variants.  Thus, the logarithm of the number of the respective halfspaces is bounded by $O(\gamma^{-2}s\log^2(\beta^{-1})).$ This is a bound on the VC dimension of the class of halfspaces that potentially can be constructed by the modified LP algorithm. To obtain the VC dimension of the class of $t$-polyhedra built based on this class of halfspaces, this bound should be multiplied by $t.$ The sample complexity is then obtained by the respective replacement of $D$ in Section "PAC polyhedral classification" by the above bound on the VC dimension. For the improper-separation algorithm, we use $t^{O(\gamma^{-2})}$ in place of $t.$

\end{document}